\documentclass[10pt,twocolumn,a4]{article}

\usepackage{times}
\usepackage{epsfig}
\usepackage{graphicx}
\usepackage{amsmath}
\usepackage{amssymb}

\usepackage{subfigure}

\usepackage{booktabs}
\usepackage{multirow} 

\usepackage[pagebackref=true,breaklinks=true,letterpaper=true,colorlinks,bookmarks=false]{hyperref}

 \author{\textbf{Gautam S. Thakur, Mohsen Ali,  Pan Hui and Ahmed Helmy} \\ The Department of Computer and Information Science and Engineering \\ University of Florida, Gainesville \\ Email: \textit{\{moali, gsthakur\}@cise.ufl.edu, pan.hui@cl.cam.ac.uk, helmy@cise.ufl.edu}}
\begin{document}
\date{} 

\title{Comparing   Background Subtraction Algorithms and \\ Method of Car Counting }

\maketitle
\begin{abstract}
 In this paper, we compare various image background subtraction algorithms with the ground truth of cars counted.  We have given a sample of thousand images, which are the snap shots of current traffic as records at various intersections and highways. We have also counted an approximate number of cars that are visible in these images. In order to ascertain the accuracy of algorithms to be used for the processing of million images, we compare them on many metrics that includes (i) Scalability (ii) Accuracy (iii) Processing time. 
  
\end{abstract}

\section{Introduction}
There are thousands, if not millions, of outdoor cameras currently connected to the
Internet, which are placed by governments, companies, conservation societies, national
parks, universities, and private citizens. We view the connected global network
of webcams as a highly versatile platform, enabling an untapped potential to monitor
global trends, or changes in the flow of the city, and providing large-scale data to realistically model vehicular, or even human mobility. We developed a crawler that collects vehicular mobility traces from these online webcams. A majority of these webcams are deployed by city's Department of Transportations (DoT). These web cameras are  installed on  traffic signal poles facing towards the roads of some prominent intersections throughout city and highways. At regular interval of time, they capture still pictures of on-going road traffic and send them in the form of feeds to the DoTs media server. For the purpose of this study, we made agreements with DoTs of 10 cities with large   coverage    to collect these vehicular imagery data for several months. We cover cities in North America, Europe, Asia, and Australia. 

\begin{table} 
\centering
\label{dataset}
\scalebox{.7}{
\begin{tabular}{|c| c| c| c|} \hline
\textbf{City} & \textbf{ \# of Cameras} & \textbf{Duration} &  \textbf{  Records}  \\ \hline
Bangalore & 160 & 30/Nov/10 - 01/Mar/11 & 2.8 million  \\\hline
Beaufort & 70 & 30/Nov/10 - 01/Mar/11 & 24.2 million  \\  \hline
Connecticut & 120 & 21/Nov/10- 20/Jan/11 &   7.2 million  \\ \hline
Georgia & 777 & 30/Nov/10 - 02/Feb/11 &  32 million  \\  \hline
London & 182 & 11/Oct/10 - 22/Nov/10 &  1 million \\ \hline
London(BBC) & 723 & 30/Nov/10 - 01/Mar/11 &  20 million \\ \hline
New york & 160 & 20/Oct/10 - 13/Jan/11 &  26 million  \\  \hline
Seattle & 121 & 30/Nov/10 - 01/Mar/11 &  8.2 million  \\  \hline
Sydney & 67 & 11/Oct/10 - 05/Dec/10 & 2.0 million  \\  \hline
Toronto & 89 & 21/Nov/10 - 20/Jan/11 & 1.8 million  \\  \hline
Washington & 240 & 30/Nov/10 - 01/Mar/11 &  5 million  \\\hline
\textbf{Total}	&  \textbf{2709} &\textbf{-}  & \textbf{125.2 million} \\
\hline\end{tabular}}\vspace{-2.5mm}\caption{{Global Webcam Datasets}}
\end{table}

\begin{figure}
\centering
\mbox{\subfigure[London]{
   \includegraphics [ scale=.2]{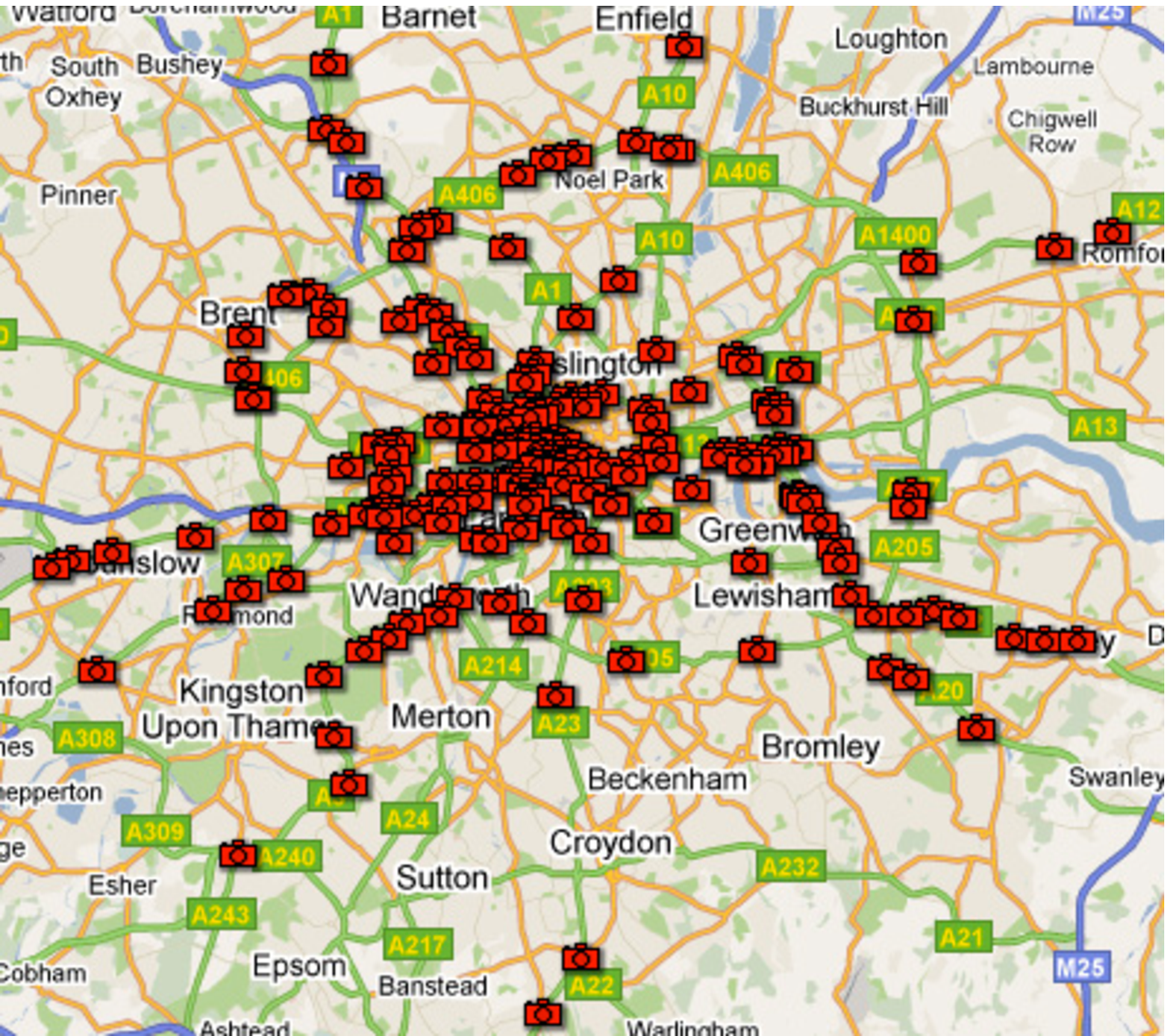}
 }
\quad
\subfigure[Sydney]{
   \includegraphics[ scale=.2] {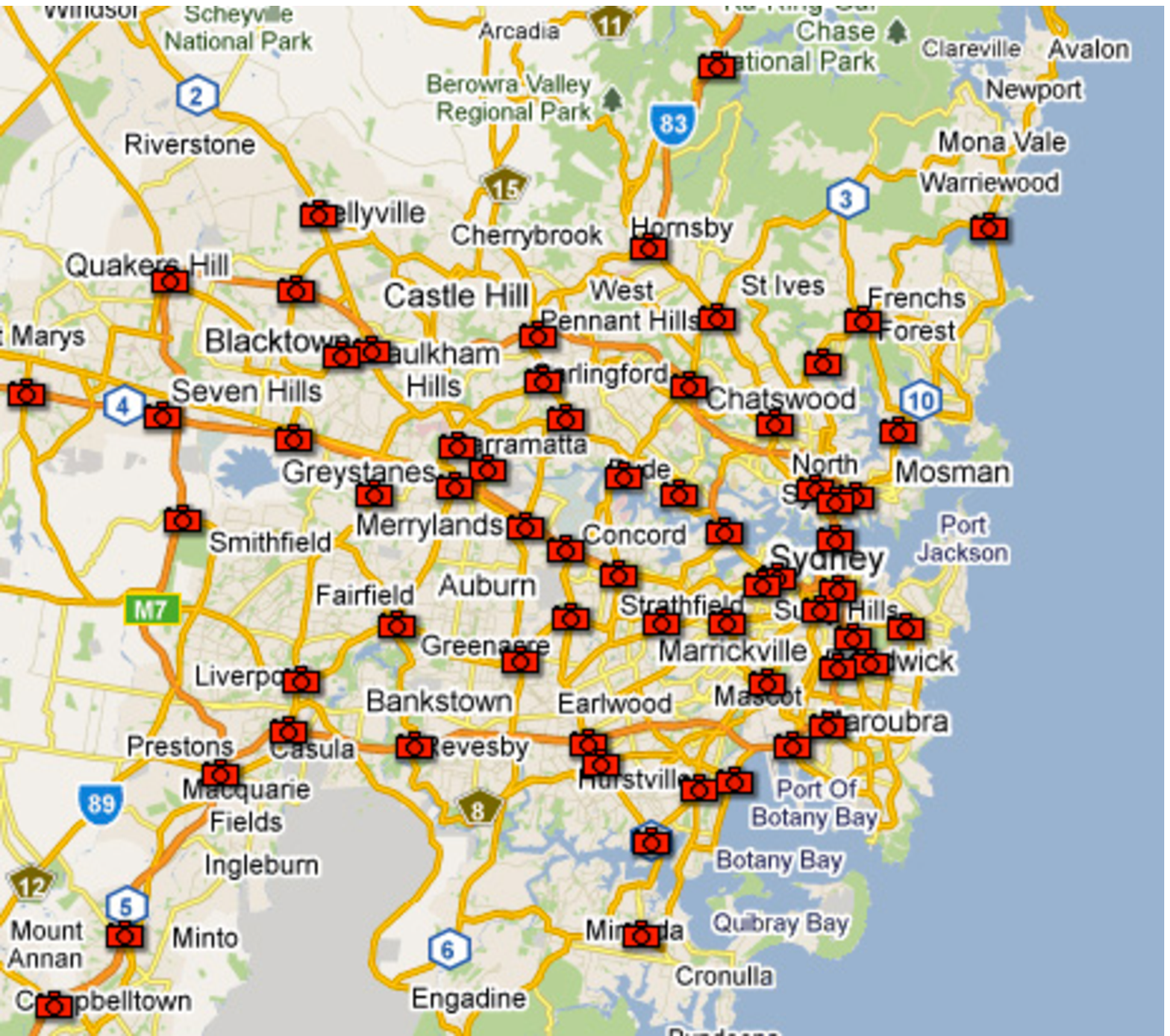}
 }
  }\vspace{-4.5mm}
\caption{The red dots show the location of cameras  in London and Sydney.}
\label{citymaps}
\end{figure}

Since, these cameras provide better imagery during the daytime, we limit our study to download and analyze them only during such hours. On average, we download 15 Gigabytes of imagery data per day from over 2700 traffic web cameras, with a overall dataset of 7.5 Terabytes  containing around 125 million images. Table-\ref{dataset} gives a high level statistics of the dataset. Each city has a different number of deployed cameras and a different interval time to  capture images. In Fig.-\ref{citymaps}, we show a geological snapshot of the cameras deployed in the city of London and Sydney. The area covered by the cameras in London is $950km^2$ and that in Sydney is $1500km^2$. Hence, we believe our study will be comprehensive and will reflect major trends in traffic movement.        Next, we discuss the algorithm to extract traffic information from images.

\section{Background Subtraction}
\label{sec:BS}
Background subtraction is a standard method for the object localization in the video sequences especially for the 
surveilling applications where cameras are fixed. 
In the environment and applications greatly where simple object detection
is not possible (because object could not be modeled due to variations) or is too expansive, background-subtraction methods
are used to remove regions that might not be object.
In most of the surveilling videos where cameras  are static that is the "background" does not change much 
 (in comparison to foreground objects)
across the time and thus could be modeled. Any object or part of image that does not follow that 
model is characterized as "foreground". These "foreground" regions are then 
further processed to analyze if they represent desired object or not. 

The straight forward technique of background subtraction is to just subtract previous frame with the current frame and threshold
the result on each pixel. But such a straightforward method fails when the object moves very slowly. Evolving from this 
is to represent a single handpicked image that does not have any "foreground" object, as model of background. However
such an image might be difficult to obtain and will not model small variations in the background itself. 
This lead to learning background as Gaussian model for each pixel with subsequent work on how to update such model. 
See \cite{benezeth2008review} and \cite{sheikh2005bayesian} for detailed a review of background subtraction methods. 
We choose to use \cite{stauffer1999adaptive} because of it's simplicity in implementataion and ......(reliability needs
 reference)
In \cite{stauffer1999adaptive} each pixels is modeled by multiple Gaussian distributions. Let $x_t$ represent a pixel value in 
the $t^{th}$ frame, then probability of observing this value is given $K$ Gaussian distributions is 
\begin{equation}\label{eq:MOGprb}
  P(x_t) = \sum_{i=1}^{K} w_i^t*\mathcal{N}(x_t, \mu_{i,t},\Sigma_{i,t})  
\end{equation}
\begin{equation}\label{eq:MOGprbNormal}
  \mathcal{N}(x_t, \mu_{i,t},\Sigma_{i,t}) = \frac{1}{(2\pi)^{\frac{n}{2}}\|\Sigma_{i,t}\|^{\frac{1}{2}}}
    e^{-\frac{1}{2}(x_t-\mu_{i,t})^{T}\Sigma{-1}(x_t-\mu_{i,t})}
\end{equation}
where $ \mu_{i,t}$ and $\Sigma_{i,t}$ are the mean and co-variance matrix of the $i^th$ distribution. $ w_i^t$
controls how much each distribution is important. As \cite{stauffer1999adaptive} we assume the $RGB$ channels are 
uncorrelated thus the covariance matrix is diagonal. This model is updated for each image, see  \cite{stauffer1999adaptive}
for details. 

The resultant binary map obtained after the background subtraction is sent for morphological operations
 to remove the noise and refine the map by removing the blobs which have area smaller than some threshold. The 
$true$ values in the resultant binary map represents the foreground. 

Due to the perspective properties of images, a vehicle will appear smaller (that is it will use less amount of pixels
) when it's far away from camera, whereas same vehicle will appear much bigger when it's in front of camera. 
To counter this we weigh each pixel with increasing weights from bottom of the image to the top. This is based
on assuming that cameras are always upright and facing the road, therefore the car that is far from the camera
will appear on the top of image and one that is near will appear below the center of image. 

\begin{figure*}[]
\centering
\mbox{
\subfigure[Camera 5]{
   \includegraphics [width=2in]{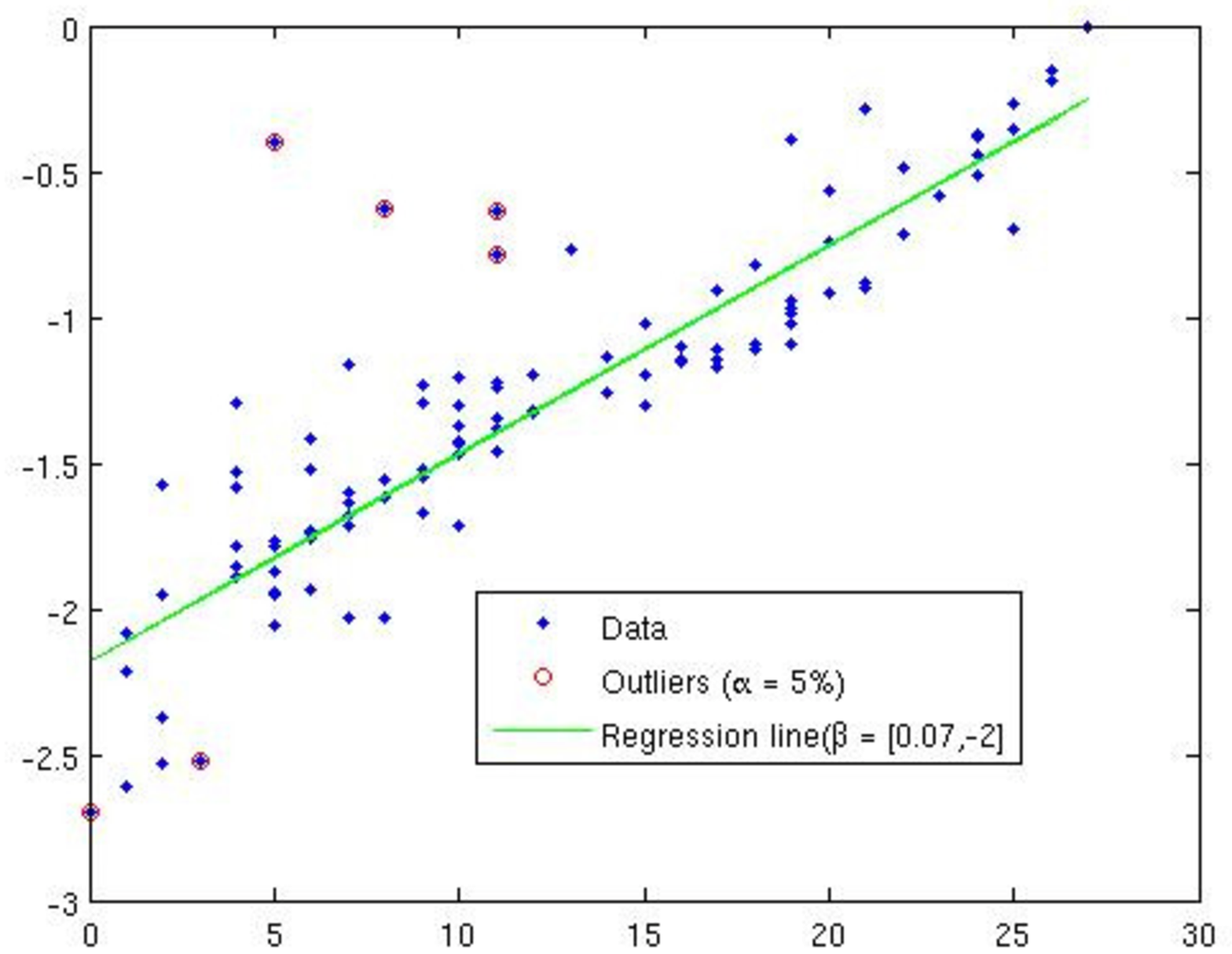}
 }
 \subfigure[Camera 11]{
   \includegraphics[width=2in] {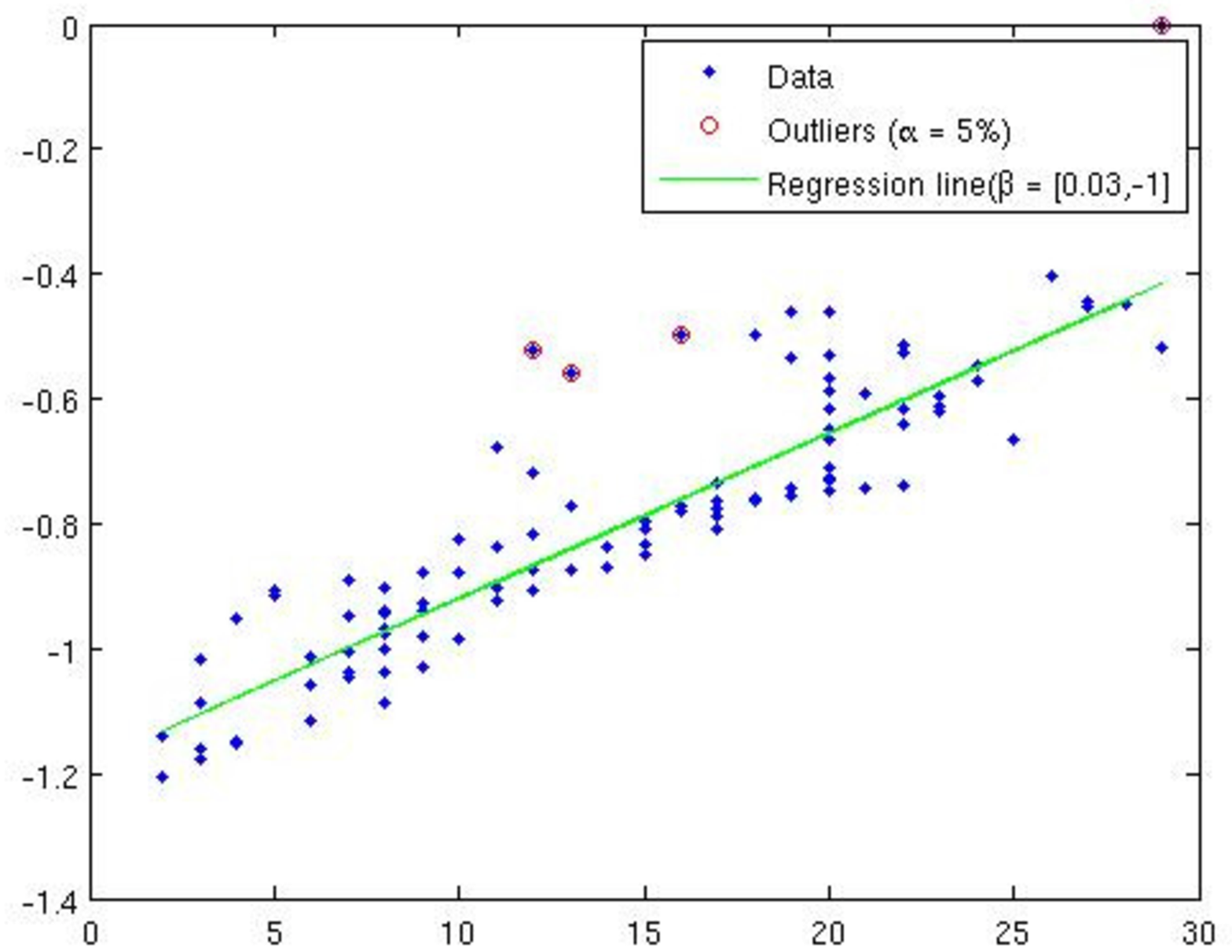}
}
\subfigure[Camera 15]{
   \includegraphics[width=2in] {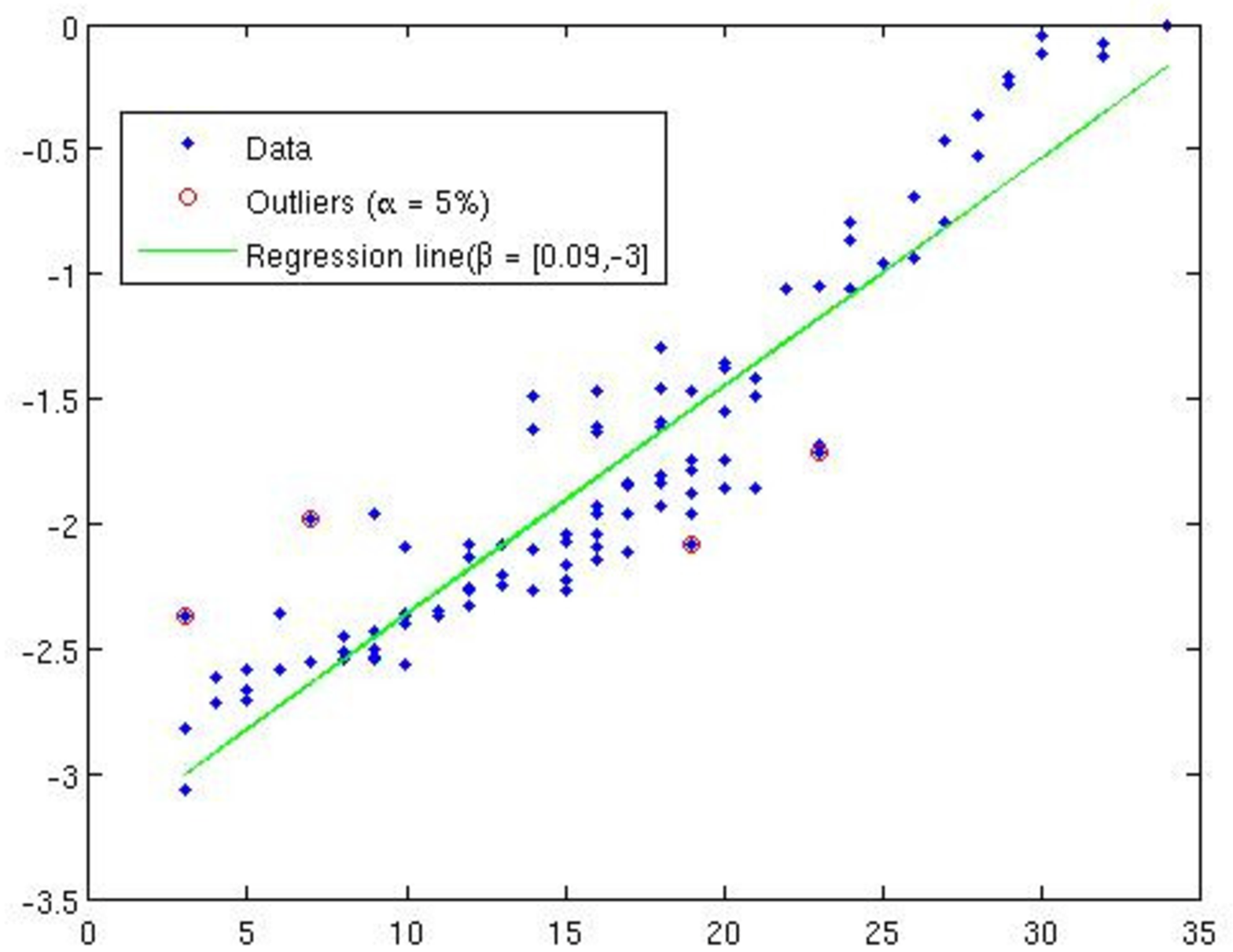}
}}
\label{figs3}
\caption{A comparison of empirical traffic densities with number of cars recorded. }
\end{figure*}

\subsection{Challenges}
\begin{itemize}
 \item Cameras are at very low angle, thus most of the times cars are occluded by other cars. Thus object detection if used
will only detects object that just in front.  
\item many times an object does not even appear to be vehicle, e.g. a when only side part of truck is visible
there are no features except a big rectangular box, it should be noted a buildings are also rectangular
\item Cameras are at different angles and locations, therefore information transfer between them is not trivial. 
 \item Low frame rate makes it impossible to use any motion based techniques like optical flow or more powerful techniques
like tracking of object or features. Many times a car or bus is visible to only 2 or three frames, therefore it is 
required to be counted even when it is very far from camera. 
\item sheer size of data makes it hard to process in timely fashion. 
\end{itemize}

\subsection{Problems and insights}
\begin{itemize}
 \item Slight camera motion due to environmental factors like air or unintentional movement by human
 \item Modeling the different parts of the day and weather
 \item Modeling and detection of road itself
 \item Modeling buildings and different natural structures for detection, using texture based features. 
 \item More region based background subtraction rather than just pixel based approach 
because as indicated in \cite{sheikh2005bayesian} and  
 \item Although the optical flow is not proper to represent the foreground but it could still be used to model
the motion in the background. 
\item Because the images are time stamped we can make a probabilistic model to represent background at different times of day.

\end{itemize}

\section{Results}
In this section, we discuss how our algorithm performs with respect to ground truths recorded. In our case, ground truths are the number of cars that are visible and handpicked for the correlation purposes.  The results are shown in the Fig.\ref{figs3}.

\bibliography{backgroundSubtraction}

\end{document}